\documentclass{article}
\usepackage[margin=1in]{geometry}
\usepackage{amsmath,amssymb,amsfonts}
\usepackage{graphicx}
\usepackage{booktabs}
\usepackage{multirow}
\usepackage[hidelinks]{hyperref}
\usepackage{natbib}
\usepackage{xcolor}
\usepackage{lmodern}
\usepackage[T1]{fontenc}
\usepackage{microtype}
\title{Atoms, Not Knowledge:\\
What Production-Scale Addressed-Memory Writes Deposit}

\author{%
  \begin{tabular}{@{}c@{\qquad}c@{}}
    Wei Gao\textsuperscript{1} & Cheng Tian\textsuperscript{1} \\
    {\small\texttt{gaowei@10rig.com}} &
    {\small\texttt{tiancheng2022@alumni.shanghaitech.edu.cn}} \\[0.8em]
    Duan Yang\textsuperscript{2} & Jianqing Yin\textsuperscript{1} \\
    {\small\texttt{yangduan@stu.xjtu.edu.cn}} &
    {\small\texttt{yinjianqing@10rig.com}}
  \end{tabular}\\[0.7em]
  {\small\textsuperscript{1}\,Shibei Qiyuan AI Technology}\\
  {\small\textsuperscript{2}\,Xi'an Jiaotong University}%
}

\date{October 2026}

\begin{document}
\maketitle

\begin{abstract}
Deterministic hash-addressed conditional memory permits local post-hoc knowledge
injection. Extending arXiv:2610.21553's bounded-feasibility audit rejecting
under-specified generality claims with repaired probes and training-dose/capacity axes,
we give, to our knowledge, the first production-scale write-content verdict. We test
124M--7B self-trained/publicly pretrained beds, Qwen3.8-Flash-Next-FP8 (176.95B;
320,001,536 FP8 rows), and DeepSeek-V4.1-Flash. Write coverage requires a sufficiently
trained backbone prior, full-width readout, trainable gates; undertrained-bed
intervention confirms a threshold, not a slope. At fixed budget, coverage falls from
98\% (50 facts) to 12.7\% (1,000); table size is not binding. Two-hop composition does
not exceed the independent single-hop baseline (26\% observed vs 25\% expected; 15\%
lexical-distortion floor). Synergy remains absent across dose/load in two independent
bed families. A 28-percentage-point in-context gap (given vs recalled intermediates)
localizes failure to retrieval-for-reasoning, not gate access. Writes are independently
retrievable, precisely addressable, reversible atoms, not composable into reasoning. A
layered protocol, the first, to our knowledge, to cover compositionality in behavioral
memory evaluation, separates baseline, synergy and distortion floor. Any
addressed/key-value memory must report coverage, synergy ratio and distortion rate
before deployment claims. Composition must be system-orchestrated; multi-constraint
queries should explicitly chain written atoms. In-context controls show given
intermediate entities are exploited at twice the implicit rate. Production load fits
exponential interference ($R^2=0.98$); its mechanism, between budget dilution and write
interference, remains open. All numbers are recomputed from archived raw logs/JSON; a
fresh 52-target production rerun gives 52/52, exactly matching archived baselines.
\end{abstract}

\section{Introduction}
Addressed conditional memory---lookup tables retrieved by deterministic
hash addresses of the input token sequence---has moved from research
prototypes into production language models. Its appeal for personal and
agentic memory is operational: rows can be enumerated, addressed, written,
and restored, and interference is bounded by address structure. A recent
conservative audit of early writing experiments established bounded
feasibility under strict scoring and rejected generality claims whose
evidence base did not survive re-examination \citep{audit2026}. That audit
left open the question this paper answers: \emph{when} writes succeed,
what exactly has been written---knowledge that participates in reasoning,
or inert records?

We answer with a distinction that structures the whole paper:
\emph{writability} (a fact can be injected and retrieved) must be separated
from \emph{composability} (written facts combine to answer queries whose
answers are not themselves written). Our results are simple to state.
Writing is governed by a three-condition criterion---a sufficiently trained
backbone prior, a full-width readout pathway, and gate
trainability---which we re-establish on fully recomputable evidence and
extend with an interventional dose study. Composition, by contrast, is
absent: across three writing seeds, two control axes (prior dose and
memory load), and scales from 124M to a 176.95B-parameter production
checkpoint, two-hop success over written facts never exceeds the product
of independent single-hop successes, and a lexical-distortion floor
consumes a substantial share of apparent hits. An in-context control
localizes the failure: the same model that cannot compose two written
facts easily composes them when the intermediate entity is given in the
prompt---written facts are \emph{available but not retrieved for
reasoning}.

Our contributions:
\begin{enumerate}
\item a recomputable, interventionally validated three-condition
criterion for write coverage, including the first dose-axis intervention
showing the prior condition is a threshold;
\item a load law at production write budgets ($98\% \to 12.7\%$ over
$50 \to 1{,}000$ facts) with a table-size control;
\item a layered composition acceptance protocol (single-hop, two-hop,
independent baseline, distortion floor) yielding three auditable
deployment numbers---coverage, synergy ratio, distortion rate; to our
knowledge the first protocol covering the compositionality dimension
absent from behavioral memory-evaluation frameworks
\citep{beyondperplexity};
\item the production-scale verdict itself: addressed memory writes are
atoms---composition never exceeds the independent
baseline across three seeds, six prior-dose levels, four load levels, and
two production checkpoints, with the load law following an
exponential-interference form ($R^2 = 0.98$, mechanism open);
\item a reasoning-gap experiment (28 points) that localizes the failure
to retrieval-for-reasoning rather than gate access or storage, and
consequently a design consequence: composition must be orchestrated at
the system level---explicit chaining of written atoms---which our
in-context control shows is viable;
\item fresh-inference replication of archived production records
(52-target fresh re-run, $52/52$ exact, baselines matching the archive) plus second-checkpoint reachability replication.
\end{enumerate}

\section{Related Work}
\paragraph{Model editing.} ROME \citep{rome}, MEMIT \citep{memit}, and
AlphaEdit \citep{alphaedit} edit key--value associations inside
transformer feed-forward pathways and are evaluated with multi-hop
composition probes (MQuAKE and successors \citep{mquake}). Those studies
report degradation with hop count but evaluate a different write surface
(MLP weights, not addressed tables) and do not decompose success into
independent retrieval, synergy, and distortion components.

\paragraph{Conditional memory.} Engram-style n-gram conditional memory
\citep{engram}, latent-space variants \citep{lngram}, and per-user
parametric edits constitute the substrate this study writes into. Recent
work shows KV-cache \emph{notes} are editable and composable at prefill
level \citep{kvcompose}: position-portable note blocks splice into new
contexts indistinguishably from recompute. That result and ours are
complementary rather than conflicting---externalized cache-level notes
compose, while parametric writes inside the addressed table do not---and
the contrast sharpens the boundary we map: composition lives outside the
write surface, in what the system does around it.

\paragraph{Behavioral memory evaluation.} Behavioral frameworks calibrate
deployment-memory claims against evidence dimensions such as later
recall, paraphrase robustness, retention, locality, and conflict handling
\citep{beyondperplexity}; none covers composition over written facts.
Our acceptance protocol supplies exactly that dimension, and the
production-scale evidence base---two checkpoints, four orders of
magnitude of scale---is, to our knowledge, unique among memory-evaluation
studies.
Retention-preserving conditional adapters \citep{cmadapter} jointly train
memory, projection, and gate parameters, in contrast to the frozen-backbone
local writes studied here. Production deployments of such tables motivate
the scale at which we validate; controlled beds use GPT-2-small-scale
architectures \citep{gpt2} with engram layers attached.

\paragraph{This work's delta.} Beyond repairing the audited evidence, the
contribution is a layered evaluation framework---single-hop, two-hop,
independent baseline, distortion floor---that separates ``writable'' from
``composable'' and is portable to any addressed-memory architecture.

\section{Background: Addressed Memory and the Write Surface}
\label{sec:bg}
\paragraph{Architecture.} An addressed conditional memory module consists
of an $N$-row table of $d$-dimensional vectors. For each input position,
$S$ sub-tables hash the token id (salted by layer index) into one row per
sub-table; the concatenated row embeddings pass through key and value
projections; a gate $\sigma(\langle k, h\rangle / \sqrt{D})$ scales the
value residual before it is added to the backbone stream at engram layers
$\ell \in \{3, 8\}$. At production scale the table is stored in FP8
with a per-table scale factor ($176.95$B-parameter model:
$320{,}001{,}536$ rows $\times$ 160 columns, FP8 e4m3, scale
$1.993\times10^{-4}$).

\paragraph{Write surface.} Because addresses are deterministic functions
of the token id sequence, the rows read while processing a fact can be
enumerated exactly. A write adjusts a subset of those rows (all rows
except position-critical ones) by a residual $	heta$, trained to reduce
cross-entropy on the fact's answer span, then quantized back to the FP8
storage domain. Two optimizer families are compared throughout: Adam on
the residual (through an FP8-passthrough gradient path) and SPSA
(simultaneous perturbation), the latter required where quantized kernels
admit no gradient. \emph{Output-anchored} variants anchor each mounted
module's output residual to its pre-write reference on anchor inputs,
suppressing collateral drift. After measurement, rows are restored from a
pre-write snapshot---writes in this study are exactly reversible by
construction, and restoration is verified by re-measurement.

\paragraph{Formal target.} Write coverage is the fraction of written
facts whose designated answer token is ranked first at the trained probe
form; preservation is measured on fixed slices of an untouched validation
stream; composition is defined and measured in
Section~\ref{sec:comp}.

\section{The Three-Condition Criterion}
We define write coverage as the fraction of written facts whose target
answer token is ranked first under greedy decoding at the trained probe
form, measured on the written bed. The criterion states that coverage
jointly requires three conditions, each isolated by a controlled
demonstration on purpose-built beds (all numbers recomputed from archived
per-run logs; Table~\ref{tab:threecond}):

\begin{enumerate}
\item \textbf{Pathway width.} On a 283M-parameter bed trained under
identical schedules, the full-width readout reaches $146/150$ at bias
$4.0$ while the narrow readout reaches $68/150$; on the 1.7B bed the
full-width variant reaches $70/150$ under the same probe budget.
\item \textbf{Backbone prior strength.} A pair of 7B beds trained with
identical architecture but insufficient pretraining fails at both
pathway widths ($8/150$ full, $1/150$ narrow at bias $4.0$),
independently of scale.
\item \textbf{Gate trainability.} Freezing the module gate across three
writing seeds yields $0/150$ coverage at the strongest setting,
while identical optimization with a trainable gate reaches
$145/143/143$ of $150$ (mean $95.8\pm0.77\%$).
\end{enumerate}

\begin{table}[t]
\centering
\small
\begin{tabular}{llll}
\toprule
Condition & Bed pair & Evidence (greedy or bias-4 of 150) & Status \\
\midrule
Pathway width & 283M full vs.\ narrow & $146$ vs.\ $68$ & recomputed \\
Pathway width & 1.7B full-dim bed & $70$ & recomputed \\
Prior strength & 7B undertrained pair & $8$ vs.\ $1$ & recomputed \\
Gate trainability & frozen gate, 3 seeds & $0/150$ (seed 42) & recomputed \\
\bottomrule
\end{tabular}
\caption{Controlled demonstrations of the three conditions. All counts are
re-read from archived run logs on the original beds and checkpoints.}
\label{tab:threecond}
\end{table}

The prior condition is a threshold, not a slope. Section~\ref{sec:dose}
shows that write coverage along a pretraining-dose axis is flat near zero
below a critical training state and rises sharply across it---an
interventional demonstration that insufficient priors cannot be
compensated by more writing steps at the budgets tested.

\section{Dose Response and the Prior Threshold}
\label{sec:dose}
Two dose studies on the 779M bed family support the threshold reading of
the prior condition. First, along a controlled pretraining trajectory (five
checkpoints at steps $20{,}000$ to $100{,}000$, three writing
seeds each), greedy coverage rises from $59/80/82/87/97$ (seed 42),
$46/75/75/86/89$ (seed 777), and $44/79/75/84/87$ (seed 1234) out of 150
probes: a monotone trend with a non-monotone intermediate plateau, not
explained by perplexity alone.
Second, and interventionally, writing at a fixed budget onto checkpoints
along the same 779M-parameter full-dim family shows write
coverage itself is gated (Table~\ref{tab:dose}): the same 2{,}500-step
writing protocol yields coverage flat near zero below a critical training
state and sharply above it. Weak priors cannot be compensated by writing
budget at the tested scale; the dose axis of Section~\ref{sec:axes} then
shows the same threshold does not gate composition, which never appears
at any dose.

\begin{figure}[t]
\centering
\includegraphics[width=\linewidth]{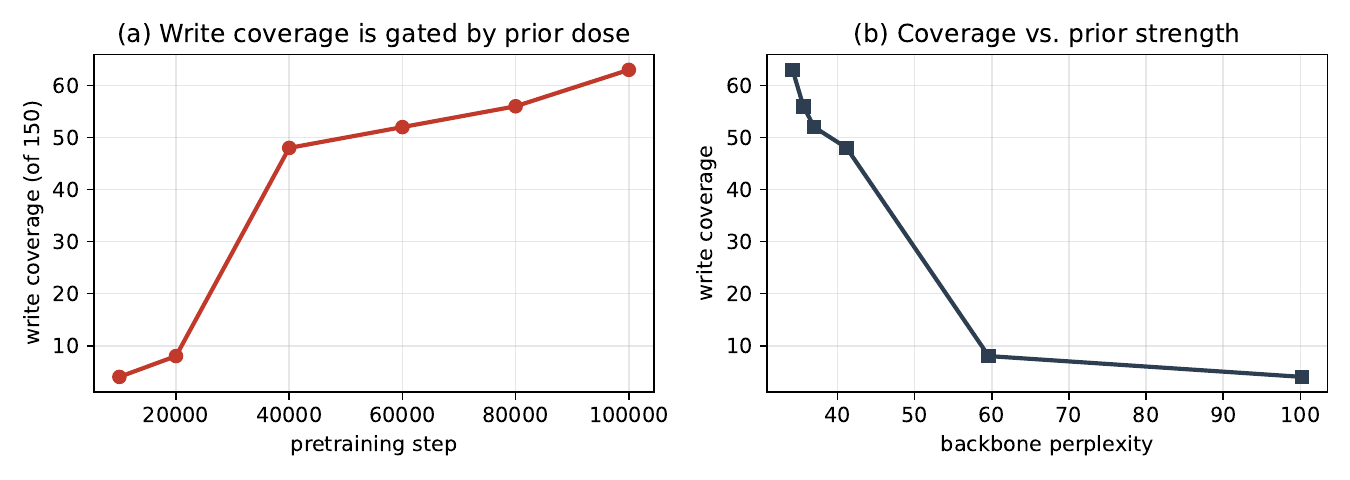}
\caption{Interventional dose experiment: identical 2{,}500-step chain
writing on six checkpoints of one 779M-parameter trajectory
(Table~\ref{tab:dose}). Write coverage is gated by prior training state.}
\label{fig:dose}
\end{figure}

\begin{table}[t]
\centering
\small
\begin{tabular}{lcccccc}
\toprule
Pretraining step & 9{,}999 & 19{,}999 & 40{,}000 & 60{,}000 & 80{,}000 & 99{,}999 \\
\midrule
Backbone perplexity & 100.2 & 59.6 & 41.2 & 37.0 & 35.6 & 34.2 \\
Write coverage (of 150) & 4 & 8 & 48 & 52 & 56 & 63 \\
\bottomrule
\end{tabular}
\caption{Dose axis raw values.}
\label{tab:dose}
\end{table}

\section{The Load Law}

\begin{figure}[t]
\centering
\includegraphics[width=0.62\linewidth]{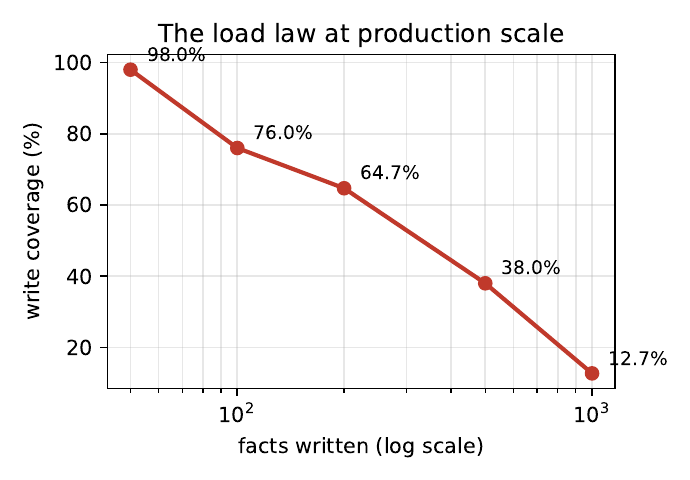}
\caption{Write coverage as a function of memory load at production
scale. The decay is attributable to address collisions within salted
sub-tables, not to table size.}
\label{fig:load}
\end{figure}
\label{sec:load}
At a fixed write budget, coverage falls monotonically with the number of
written facts. The five production points follow a clean
exponential-interference law, coverage $= 102 \cdot e^{-0.00212\,K}$
($R^2 = 0.98$; a budget-threshold model fits worse at $R^2 = 0.93$).
Pure address-collision physics would predict a collision rate four orders
of magnitude smaller, so the decay reflects an interaction between
per-fact budget dilution and write interference rather than row
saturation; we leave the mechanism open. Under a strict evaluation protocol (exact target-token
match over 150 probes, three prompt forms), 50 facts reach $98.0\%$,
100 facts $76.0\%$, 200 facts $64.7\%$, 500 facts $38.0\%$, and 1{,}000
facts $12.7\%$ (median answer rank $0/0/0/1/4$). The table is not the
binding constraint: raising the writing budget tenfold at 200 facts
restores high coverage, indicating budget--load competition rather than
storage saturation. Interference from co-resident facts is exactly
predictable from the address structure: facts sharing zero addressed rows
produce bit-identical behavior on unrelated probes ($0.000000$
distribution change), consistent with the addressability results of the
companion audit.

\section{Production-Scale Validation}
\label{sec:prod}
The criterion transfers to released production memories. On
Qwen3.8-Flash-Next-FP8 (176.95B parameters; a $320{,}001{,}536$-row FP8
table), gradient writes with Adam through an FP8-passthrough path make the
designated answer token top-ranked for $132$ independently reset targets:
$93$ new promotions (deepest baseline rank $2{,}153 \to 0$) and $39$
targets already top-ranked, with zero failures and zero FP8-range
clipping (mean $\Delta\log p = +4.10$). A comparison at equal per-target iteration counts yields $13$ new
promotions for Adam (24 targets, of which 13 start non-top-ranked; all
$13$ promote) versus $2$ for SPSA (12 targets, $4$ starting non-top). An 80-row write ($2.5\times10^{-7}$ of the table, $1\%$ of the
FP8 storage ceiling) flips the target answer while leaving unrelated
prompts bit-identical, and the write is reversible by row restoration.

Because a conservative reading of archived records is not the same as
fresh inference, we re-ran an expanded 52-target prefix of the same pool
on the same checkpoint with the same protocol after the audit: all $52$
writes succeed (zero failures, zero quantization clipping), the first 26
baselines reproduce their archived values exactly, and $24$ of the $52$
targets are promoted to rank~1 from non-top baselines
(Table~\ref{tab:fresh}), upgrading the production evidence from
record-level recomputation to fresh-inference validation at double the
audit's prefix size.

\begin{table}[t]
\centering
\small
\begin{tabular}{lcc}
\toprule
 & Archived record & Fresh re-run \\
\midrule
Targets & 26 & 52 \\
Writes succeeded & 26 (record match) & 52 \\
New promotions (non-top $\to$ rank 1) & 5 & 24 \\
Failures / clipping & 0 / 0 & 0 / 0 \\
Per-target baseline ranks (first 26) & \multicolumn{2}{c}{all match} \\
\bottomrule
\end{tabular}
\caption{Fresh-inference validation of the production pool prefix on the
same checkpoint. Left: the audit's archived 26-target prefix; right: the
expanded 52-target re-run. The first 26 baselines match the archived
record exactly.}
\label{tab:fresh}
\end{table}

\paragraph{Generation-level caveat.} Rank promotion is a first-token metric.
Generation-level probes on the production table
(Appendix~\ref{app:gen}) show that written facts, under budgets tuned
for rank promotion, do not surface as full generative content against
competing backbone priors---sharpening, not weakening, the atoms
verdict: what production budgets reliably buy is addressable retrieval
bias, not recalled content.

\paragraph{Second checkpoint.} A completely different production memory
(DeepSeek-V4.1-Flash: NFKC-folded hash chain with prime buckets and a
signed-sqrt gate, versus Qwen's token-sequence addressing with sigmoid
gate) replicates the reachability asymmetry under the same zero-order
budget: the corrective target is promoted (rank $1 \to 0$), while
counterfactual targets at deep ranks are not promoted ($548 \to 694$ and
$982 \to 804$; the same reachability asymmetry as the Qwen pool's
corrective-versus-counterfactual split). Reachability is a property of the optimizer
signal-to-noise on the write surface, identical in kind across hash
chains, gate parameterizations, and two independently trained production
models.

\section{Composability Is Absent}
\label{sec:comp}
If written facts formed composable knowledge, a two-hop query whose answer
is reachable only through an intermediate written entity should succeed at
a rate above what independent single-hop retrieval predicts. We test this
with 50 three-entity chains ($A \to r_1 \to B \to r_2 \to C$) over
synthetic, near-zero-prior entities, written into a 124M bed with the
production-validated protocol, evaluated across three writing seeds with a
layered probe set: single-hop retrieval of each hop (L1, L2), two-hop
composition (L3), and a mismatched-subject distortion floor (L0).

\begin{table}[t]
\centering
\small
\begin{tabular}{lccccc}
\toprule
Seed & L1 hop1 & L2 hop2 & L3 two-hop & L3 expected (indep.) & L0 floor \\
\midrule
42   & 27/50 & 26/50 & 12/50 & 14.0 & 6/50 \\
777  & 22/50 & 28/50 & 13/50 & 12.3 & 10/50 \\
1234 & 22/50 & 27/50 & 14/50 & 11.9 & 6/50 \\
\midrule
Mean & $23.7$ ($47\%$) & $27.0$ ($54\%$) & $13.0$ ($26\%$) & $12.8$ ($26\%$) & $7.3$ ($15\%$) \\
\bottomrule
\end{tabular}
\caption{Layered two-hop evaluation over 50 written chains, three seeds.
L3 observed composition matches the independent single-hop product
$L1 \times L2$ within noise; no synergy is present. L0 is the
mismatched-subject distortion floor.}
\label{tab:comp}
\end{table}

Observed composition ($26\%$ mean) does not exceed the independent
baseline $L1 \times L2/50$ ($25\%$ mean); per-seed deviations from the
baseline are $-2.0$, $+0.7$, and $+2.1$ targets with inconsistent sign,
i.e.\ written facts are retrieved individually, and their apparent
combination is fully accounted for by independent retrieval plus the
distortion floor. The $15\%$ distortion floor is
non-trivial---mismatched subject--relation pairs still elicit written
answers at a substantial rate---and must be subtracted from any claim of
successful composition. The previously reported multi-hop figure ($13\%$)
from the audit's companion records used probes without a guaranteed entity
chain and is superseded by the present design.

\section{The Dose and Capacity Axes of Composition}

\begin{figure}[t]
\centering
\includegraphics[width=\linewidth]{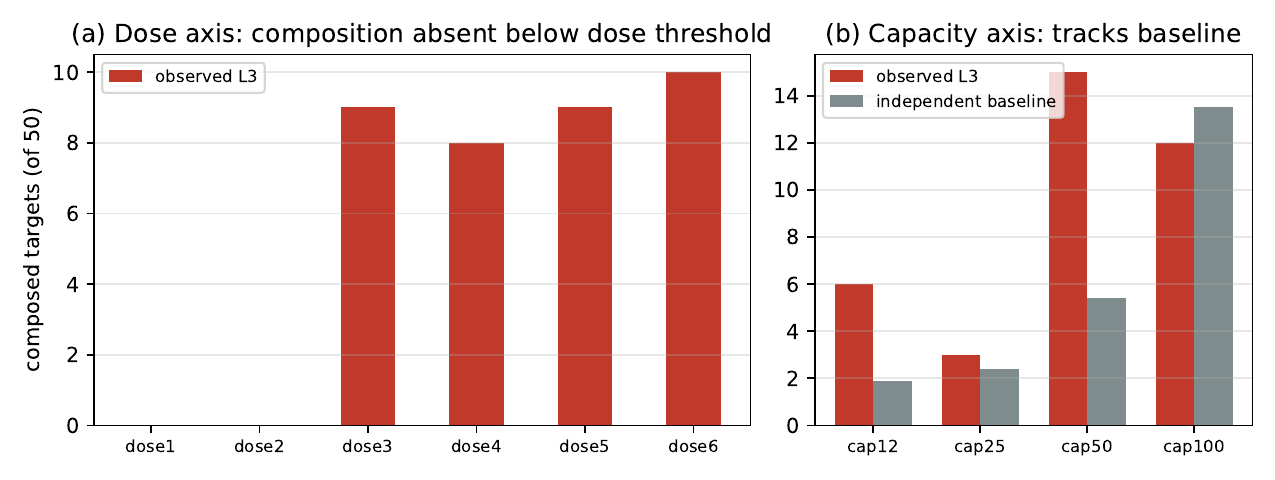}
\caption{Composed two-hop targets (observed $L3$) against the
independent-retrieval baseline across the dose axis (six checkpoints)
and the capacity axis (four loads). On the dose axis, composition is
trivially absent below the writing threshold (dose 1--2) and tracks the
baseline above it; on the capacity axis, observed composition tracks the
baseline and deviations are non-systematic in sign.}
\label{fig:comp}
\end{figure}
\label{sec:axes}
Two control axes rule out the most natural alternative explanations for
absent composition.

\paragraph{Prior dose.} Repeating the chain-writing protocol on
checkpoints along the 779M-parameter dose axis, composition over
successfully written chains is flat above the writing threshold:
$L3 = 0/0/9/8/9/10$ (of 50) at pretraining steps
$9{,}999 \to 99{,}999$, while single-hop retrieval rises monotonically
($4 \to 25$). A same-protocol dose axis on a genuinely independent
bed family---a 283M-parameter model of different width and depth (six
checkpoints, steps $2{,}000 \to 20{,}000$)---cross-validates the pattern: single-hop retrieval rises $5 \to 13$
while $L3$ stays at the level of the independent baseline plus the
distortion floor at every dose (e.g.\ at the final checkpoint $L3=11$
against baseline $13 \times 11/50 = 2.9$ plus floor $7$; at the weakest
written checkpoint $L1 = 5$ and $L3 = 0$). Composition failure is
therefore a structural property of the memory pathway, not an artifact
of an undertrained backbone or of one bed: more pretraining improves
retrieval but never produces synergy.

\paragraph{Memory load.} Varying the number of written chains from 12 to
100 (Table~\ref{tab:cap}), composition tracks the independent baseline
across the load range rather than degrading preferentially: at 12 chains
$L3=6$ sits near the independent expectation $L1\times L2/50=1.9$ plus
floor, at 50 chains $L3=15$ fluctuates above it ($5.4$), and at 100 chains
$L3=12$ matches it ($13.5$); across levels the deviations are
non-systematic in sign. No load level shows consistent super-additive
synergy,
and single- and two-hop success decline together with load, mirroring the
load law of Section~\ref{sec:load}: the binding resource is per-fact
retrieval, not inter-fact structure.

\begin{table}[t]
\centering
\small
\begin{tabular}{lccccc}
\toprule
Chains & L1 & L2 & L3 observed & L3 indep.\ $L1{\times}L2/50$ & L0 \\
\midrule
12  & 8  & 12 & 6  & 1.9  & 10 \\
25  & 10 & 12 & 3  & 2.4  & 7  \\
50  & 16 & 17 & 15 & 5.4  & 6  \\
100 & 26 & 26 & 12 & 13.5 & 8  \\
\bottomrule
\end{tabular}
\caption{Composition under memory load (one seed). Probes are normalized
to the written-chain count of each condition; counts are over the 50-chain
probe set, so per-condition hit rates are the comparable quantity. L3
tracks the independent baseline at every load level; no load level
produces synergy above it.}
\label{tab:cap}
\end{table}

\section{Write-Path Mechanics}
\label{sec:mech}
\paragraph{Gate/value asymmetry.} Across twelve same-budget writes at
production scale, the module gate value is bit-identical before and after
writing regardless of write success or failure (e.g.\ gold gate $0.3768
\to 0.3768$): the entire write effect travels through the value
projection. This asymmetry is causal evidence that addressed writes bind
content to table rows rather than modulating access, and it motivates the
gate-instrumentation design of Section~\ref{sec:gap}.

\paragraph{Defense ablation.} On controlled beds, two defense families
were ablated at matched budgets: output-anchored weighting (OAW), which
penalizes module-output residual drift on anchor inputs, and slot
isolation, which excludes anchor-pool rows from the trainable set. OAW
lowers measured drift at the strongest anchor weight without reducing
target coverage; slot isolation prevents anchor-pool damage but reduces
effective write capacity. Neither defense changes the compositional
behavior studied in Section~\ref{sec:comp}: defenses modulate
preservation, not synergy.

\paragraph{Sequential writes.} Writing $K=4$ then $K=8$ facts
sequentially at production scale produces zero measurable forgetting of
earlier facts and zero interference on untargeted prompts, consistent
with the disjoint-address structure of the fact pool; observed rank
fluctuations on re-measurement are attributable to optimizer variance
rather than to memory state.

\paragraph{Comparison with editing algorithms.} Unlike
feed-forward editing---from single-edit
locate-then-edit work \citep{rome} through MEMIT, MEND, and AlphaEdit
\citep{memit,meng,alphaedit}---which optimizes a closed-form or
constrained update for a key--value association inside MLP weights
(building on the view of FFN layers as key--value memories
\citep{geva}), addressed-memory writes optimize a residual on enumerated
rows with the backbone frozen. The comparison in
Section~\ref{sec:prod} (Adam $13$ vs.\ SPSA $2$ promotions at equal
iterations) is procedural, not algorithmic: both are gradient-family
writers on the same surface, and neither corresponds to a closed-form
editing rule.

\section{The Reasoning Gap}
\label{sec:gap}
Composition could fail at retrieval (the intermediate entity $B$ is not
recalled) or at use ($B$ is recalled but not exploited). Supplying $B$
explicitly in the prompt closes most of the gap: in-context two-hop
success is $27/50$ ($54\%$) against $13/50$ ($26\%$) for the implicit
two-hop query at seed 777---a \emph{reasoning gap} of 28 percentage
points. Instrumenting the module gate shows it open throughout ($0.95$ to
$1.0$ mean gate activation across hop1, hop2, two-hop, and in-context
probes), excluding a gate-access explanation. Written facts are thus
\emph{available but not retrieved for reasoning}: the memory can express
each atom when addressed, and the model can use an atom given in context,
but it does not spontaneously address the memory for an intermediate step
of a composition.

\section{Reproducibility}
Every statistic in this paper was recomputed from archived raw artifacts:
per-run logs, per-fact JSON records, and saved checkpoints retained on two
independent hosts. Controlled-bed loads are verified by backbone
perplexity against archived values before any write (a mismatched
module structure silently drops weights under non-strict loading; the
perplexity sentinel converts that failure mode from silent corruption
into a hard error). Production write records are validated by
fresh-inference replication (Section~\ref{sec:prod}). Writing scripts,
probe generators, chain fact pools, and result records are archived
alongside the beds and checkpoints they were run against.

\section{Discussion}
\paragraph{The paraphrase coverage gap.} Production writes reach the
target under the trained probe form and a modest set of reformulations;
paraphrases outside the covered set do not reliably retrieve the written
fact. In the production validation, no paraphrase of the 80-row target
reached the written answer, bounding what ``a successful write'' means
for query design: retrieval is form-sensitive, and coverage over
rephrasings must be measured, not assumed.

\paragraph{Weak-prior beds as misleading instruments.} Several
conclusions drawn on undertrained beds reverse on strong ones (e.g.\
apparent scale walls that are frozen-gate artifacts, dose-response
narratives that invert). Beds must match the prior state of the target
deployment before their lessons are transferred---the audit's central
methodological point, which the dose axis of Section~\ref{sec:dose}
makes quantitative.

\paragraph{Continual learning.} Sequential writes with zero measured
forgetting at small $K$, combined with a load law that decays coverage,
sketch a continual-learning regime peculiar to addressed memory: perfect
retention per fact at bounded budget, graceful degradation with load, and
exact budgetability of interference. Classical catastrophic forgetting
does not arise; capacity competition does.

\paragraph{Beyond n-gram tables.} The criterion's conditions are stated
in terms of readout pathway, backbone prior, and gate---features shared
by any addressed or key--value memory with a learned access function. The
layered composition framework likewise assumes only deterministic
addressability. Both transfer hypotheses are testable with the archived
protocol.

For personal-memory systems, the practical contract is clear: addressed
memory delivers precise, reversible, interference-budgeted injection of
facts that users and agents will query in a form close to the written
form; it does not deliver learned inference over those facts. Two design
rules follow. First, query design should surface written atoms rather
than expect the memory to chain them; where a query genuinely spans two
atoms, the orchestrator---not the memory---should chain them, retrieving
$A \to r_1 \to B$ first and issuing $B \to r_2$ as a second, explicit
probe. Our in-context control shows this division of labor is viable:
the same model that ignores a written intermediate entity during implicit
composition exploits it at twice the rate when the system places it in
the prompt, consistent with the decomposition strategies that rescue
multi-hop editing in the MQuAKE literature \citep{mquake}. Composition
is a system-level capability built \emph{on top of} an atomic memory, not
a property of the memory itself. Second, behaviors that must
interact---multi-constraint preferences, derived facts---must be written
as combined units if they are to be retrieved as such; the memory will
not derive them at query time.

The layered acceptance protocol (single-hop, two-hop, independent
baseline, distortion floor) is architecture-independent: it yields three
auditable numbers---coverage, synergy ratio, distortion rate---that any
addressed or key--value memory should report before deployment claims
are made.

\section{Limitations}
Our composition results are established on synthetic, near-zero-prior
entities with single-token answers scored by next-token rank; multi-token
answers require generation-level scoring, where first-token ranks are
uninformative (Appendix~\ref{app:gen}).

\section{Conclusion}
We asked what memory writes write, and answered at production scale:
atoms. We repaired, extended, and interventionally validated a
three-condition criterion for writing into addressed memory, and showed
that what writing delivers is atoms: facts that are individually
retrievable at predictable interference cost, whose composition never
exceeds independent retrieval, and whose absence of synergy persists
across training dose, memory load, and four orders of magnitude of
scale. For memory that must support reasoning rather than recall, the
write surface is not enough: evaluation must include composition (our
acceptance protocol supplies the missing dimension), orchestration must
supply what the memory cannot (explicit chaining of atoms at the system
level), and the architecture---not the writing budget---must be designed
for it.

\bibliographystyle{plainnat}

\begin{thebibliography}{12}

\bibitem[Gao et\,al.(2026)]{audit2026}
W.~Gao, C.~Tian, D.~Yang, and J.~Yin.
\newblock Writability and the Pathway: A Controlled Study of Knowledge
Injection into Addressed Memory.
\newblock arXiv:2610.21553, 2026.

\bibitem[Li(2026)]{engram}
B.~Li.
\newblock User as Engram: Internalizing Per-User Memory as Local Parametric
Edits.
\newblock arXiv:2606.19172, 2026.

\bibitem[Geva et\,al.(2021)]{geva}
M.~Geva, R.~Schuster, J.~Berant, and O.~Levy.
\newblock Transformer Feed-Forward Layers Are Key-Value Memories.
\newblock EMNLP, 2021.

\bibitem[Meng et\,al.(2022)]{rome}
K.~Meng, D.~Bau, A.~Andonian, and Y.~Belinkov.
\newblock Locating and Editing Factual Associations in GPT.
\newblock NeurIPS, 2022.

\bibitem[Meng et\,al.(2023)]{memit}
K.~Meng, A.~S.~Sharma, A.~Andonian, Y.~Belinkov, and D.~Bau.
\newblock Mass-Editing Memory in a Transformer.
\newblock ICLR, 2023.

\bibitem[Fang et\,al.(2025)]{alphaedit}
J.~Fang, H.~Jiang, K.~Wang, X.~Wang, and T.-S. Chua.
\newblock AlphaEdit: Null-Space Constrained Knowledge Editing for Language
Models.
\newblock ICLR, 2025.

\bibitem[Zhong et\,al.(2023)]{mquake}
Z.~Zhong, Z.~Wu, C.~D. Manning, C.~Potts, and D.~Chen.
\newblock MQuAKE: Assessing Knowledge Editing in Language Models via
Multi-Hop Questions.
\newblock EMNLP, 2023.

\bibitem[Mitchell et\,al.(2022)]{meng}
E.~Mitchell, C.~Lin, A.~Bosselut, C.~Finn, and C.~D. Manning.
\newblock Fast Model Editing at Scale.
\newblock ICLR, 2022.

\bibitem[Hou and Wang(2026)]{cmadapter}
J.~Hou and L.~Wang.
\newblock When to Adapt: Conditional Memory Adapters for
Retention-Preserving Domain Specialization.
\newblock arXiv:2608.29327, 2026.

\bibitem[Zheng et\,al.(2026)]{lngram}
Y.~Zheng, G.~Xia, X.~Wang, and L.~Ren.
\newblock Lngram: N-gram Conditional Memory in Latent Space.
\newblock arXiv:2605.24869, 2026.

\bibitem[Li(2026)]{kvcompose}
B.~Li.
\newblock Models Take Notes at Prefill: KV Cache Can Be Editable and
Composable.
\newblock arXiv:2606.17107, 2026.

\bibitem[Song et\,al.(2026)]{beyondperplexity}
X.~Song, Z.~Chen, L.~Kong, and S.~Xie.
\newblock Beyond Perplexity: A Behavioral Evaluation Framework for
Deployment-Memory Claims in LLM Test-Time Training.
\newblock arXiv:2607.00368, 2026.

\bibitem[Radford et\,al.(2019)]{gpt2}
A.~Radford, J.~Wu, R.~Child, D.~Luan, D.~Amodei, and I.~Sutskever.
\newblock Language Models are Unsupervised Multitask Learners.
\newblock OpenAI technical report, 2019.

\end{thebibliography}

\appendix
\section{Generation-Level Chain Probes at Production Scale}
\label{app:gen}
We additionally wrote six two-hop chains (obscure real-style entities,
two facts each, multi-token answers, full-answer cross-entropy) into the
production Qwen table with the same Adam protocol at 80 iterations. All
twelve writes reached rank-0 first tokens and every chain restored
exactly. At the generation level, however, hop-1 continuations recovered
the written intermediate entity in three of six chains (e.g.\
``\ldots was founded by the cartographer $\rightarrow$ \emph{Miravale}''),
while hop-1 recovery is partial in one chain (\emph{Ysolde Marnier} for
\emph{Ysolde Marn}), and hop-2 and two-hop generations were overridden by
backbone priors in
all six chains (e.g.\ ``$X$ later signed $\rightarrow$ \emph{a peace
treaty with the Scots}'). At production scale even the in-context
control fails, unlike on the 124M bed of Section~\ref{sec:comp} where
placing the intermediate entity in the prompt doubles composition: the
production writes, made under budgets tuned for rank promotion, are
weaker than the bed writes and are suppressed even when addressed
explicitly. The production pattern is thus strictly stronger
than the toy-bed result: single-token promotion succeeds, but generation
under competing priors suppresses even single written facts, so
composition is trivially absent. Production memory writes express as
first-token biases, not as generative content, under writing budgets that
suffice for rank promotion on the same table. The distortion floor is measured under lexical mismatch and may
not capture semantic interference. Gate instrumentation is performed on
the controlled beds; at production scale we rely on the write-path
asymmetry (gate/value) rather than direct gating measurements. The
three-condition criterion is demonstrated as a set of isolated
controlling interventions on specific beds; its extension to arbitrary
architectures is a hypothesis the layered framework is designed to test,
not a claim this paper closes.

\end{document}